\documentclass[letterpaper, 10 pt, conference]{ieeeconf}  

\IEEEoverridecommandlockouts                              

\usepackage{graphicx} 
\usepackage{amssymb}  
\usepackage{booktabs} 
\usepackage[table]{xcolor} 
\usepackage{caption}
\usepackage{subcaption}
\usepackage{multicol}
\usepackage{xspace}
\usepackage{multirow}
\usepackage[normalem]{ulem} 
\usepackage{tcolorbox}
\usepackage{amsmath}
\usepackage{url}
\usepackage{hyperref}
\let\labelindent\relax
\usepackage{enumitem}
\usepackage{soul}
\usepackage{pifont}   
\newcommand{\cmark}{\ding{51}}          
\newcommand{\xmark}{\ding{55}}          

\title{\LARGE \bf \methodabbrev: Latent Action Pretraining with Dedicated RGB, Depth, and Unified RGB-Depth Supervision}

\author{Manish Kumar Govind$^*$, Dominick Reilly, Pu Wang, Hieu Le, Srijan Das%
\thanks{All authors are with the Department of Computer Science, University of North Carolina at Charlotte, NC, USA.}%
\thanks{*Corresponding author: mgovind@charlotte.edu.}%
}

\newcommand{\methodabbrev}{\textsc{UniLARN}\xspace}

\definecolor{revblue}{HTML}{3182bd}
\definecolor{gaingreen}{HTML}{008000}
\definecolor{rgbbox}{HTML}{ADC9E9}   
\definecolor{depthbox}{HTML}{D3E6C9} 
\definecolor{unibox}{HTML}{E69EDD}   
\begin{document}
\bstctlcite{BSTcontrol}
\maketitle
\thispagestyle{empty}
\pagestyle{empty}


\begin{abstract}
Depth provides critical geometric supervision for robot manipulation, yet existing depth-aware policies typically inject it through fused representations or auxiliary objectives, leaving its role in action learning implicit. We introduce \methodabbrev, a latent-action pretraining framework that explicitly preserves three complementary supervisory targets from RGB-D transitions: RGB-specific, depth-specific, and unified latent actions. RGB and depth transition embeddings are first discretized through a shared codebook, then fused and re-quantized through a separate unified codebook, preserving modality-specific dynamics while modeling their shared action semantics. These three latent-action targets provide separate pseudo-action supervision during VLA pretraining, avoiding both inconsistent independent learning and premature multimodal fusion without requiring robot action labels. The pretrained VLA is subsequently fine-tuned on executable actions and requires only RGB observations and language instructions at inference. Experiments in simulation and the real-world show improvements over RGB-based latent-action pretraining, particularly on geometry-sensitive manipulation tasks. Project page : \url{https://manishgovind.github.io/unilarn/}

\end{abstract}

\section{INTRODUCTION}

Depth provides critical geometric information for robot manipulation: it determines whether a motion results in approach, contact, insertion, or collision. Yet latent-action pretraining for Vision-Language-Action (VLA) models remains largely RGB-based~\cite{chen_moto_2025,bu_univla_2025}, causing the learned action representations to be dominated by appearance changes while leaving 3D geometry implicit. Depth-aware VLAs address this limitation by incorporating geometric information through visual features, auxiliary objectives, or multimodal representations~\cite{bhat_3d-cavla_2025,yuan_depthvla_2025,li2025qdepthvlaquantizeddepthprediction,qu_spatial-vla_2025}. However, these approaches do not explicitly preserve the different roles that RGB, depth, and their interaction play in supervising the action representation. The key challenge is not simply to add depth, but to control \emph{how} depth contributes to latent-action learning. RGB captures semantic and appearance-driven dynamics, depth captures geometric dynamics, and their interaction describes the shared structure of the underlying action. Collapsing these signals into a single target can obscure their individual contributions, while learning them independently fails to capture their correspondence. 

To this end, we introduce \methodabbrev, which learns three complementary latent-action targets from every RGB-D transition: an \emph{RGB-specific} latent action, a \emph{depth-specific} latent action, and a \emph{unified} latent action. These targets provide explicit supervision for visual dynamics, geometric dynamics, and their shared action semantics, respectively. \methodabbrev exposes each of them directly during latent-action pretraining rather than forcing the policy to recover these complementary cues from a single fused representation.

\methodabbrev follows three training stages. First, it learns the three latent-action targets from RGB-D observation transitions using inverse and forward dynamics. RGB and depth transition embeddings are discretized through a shared codebook to obtain modality-specific latent actions, whose embeddings are subsequently fused and re-quantized through a second codebook to produce the unified latent action. The unified latent is further constrained to predict future observations in both modalities, ensuring that it retains both visual and geometric dynamics. 

Second, the RGB-specific, depth-specific, and unified latent actions are used as separate pseudo-action targets for VLA pretraining. \methodabbrev alternates among these targets across training batches, allowing each source of supervision to shape the learned action representation without requiring the corresponding robot action annotations. Third, the pretrained model is fine-tuned on executable robot actions. Notably, depth is required only during training , while  \methodabbrev operates solely on RGB observations and language instructions at inference. 

This separation is central to the method. Our ablations show that learning RGB and depth latents independently, or retaining only a unified latent, both underperform the proposed combination of modality-specific and unified supervision. Similarly, the proposed two-stage quantization outperforms independent quantization and single-stage fusion. These results indicate that the benefit comes not merely from adding depth, but from preserving its dedicated contribution while explicitly modeling its interaction with RGB. 

Across both simulation and real-world environments, \methodabbrev improves over RGB-based latent-action pretraining, with particularly strong gains on geometry-sensitive manipulation tasks, while requiring no depth sensing or additional depth computation at deployment. 

In summary, our contributions are:
\begin{itemize}[topsep=0pt, itemsep=2pt]

    \item We introduce \textbf{\methodabbrev}, which learns and preserves RGB-specific, depth-specific, and unified latent actions, providing dedicated supervision for visual dynamics, geometric dynamics, and their shared action semantics.

    \item We propose RGB-D Latent Action Tokenization, a \textbf{two-stage latent action learning scheme} that preserves RGB and depth-specific dynamics while also explicitly modeling the interaction between RGB and depth.

    \item We use these complementary latent actions as \textbf{pseudo-action supervision for cross-modal VLA pretraining}, enabling geometry-aware action learning without robot action annotations during pretraining and requiring only RGB at deployment.

\end{itemize}






\section{RELATED WORK}
\label{sec:relatedwork}


\subsection{Vision-Language-Action Models.}
Early works~\cite{brohan_rt-2_2023,kim_openvla_2024,open_x_embodiment_rt_x_2023,cheang_gr-2_2024,zhao2025cotvlavisualchainofthoughtreasoning,li2025llarasuperchargingrobotlearning} leverage the large-scale pretraining capabilities of VLMs to predict actions by replacing or introducing discrete action tokens into the vocabulary of the underlying language model. Alternative work \cite{black_pizero_2024,li_roboflamingo_2024, li2024cogactfoundationalvisionlanguageactionmodel,nvidia2025gr00tn1openfoundation} also builds upon VLM backbones but introduces a dedicated action expert or head to learn task-specific control policies.
Beyond direct action prediction, VLAs have been extended toward multimodal reasoning and hierarchical control. Several methods~\cite{driess2023palmeembodiedmultimodallanguage,zhou2025chatvlaunifiedmultimodalunderstanding} emphasize multimodal grounding and reasoning, while hierarchical architectures such as HAMSTER~\cite{li2025hamsterhierarchicalactionmodels} and Hi-Robot~\cite{shi2025hirobotopenendedinstruction} decompose decision making into high-level semantic planning followed by low-level execution.
While these methods demonstrate strong generalization, learning low-level control at scale often requires large collections of paired robot trajectories with action supervision, which are expensive to acquire and can limit scalability across diverse embodiments and environments.

\subsection{Latent Action Representations.} To address the limitated scalability of VLA policy networks, a line of work on latent action models has emerged. These approaches learn control dynamics in a discrete latent space, enabling VLA pretraining directly from videos without requiring explicit action supervision.
Early work such as Genie~\cite{bruce_genie_2024} and LAPO~\cite{schmidt_lapo_2024} learn discrete latent actions from game environments. Dynamo~\cite{cui2024dynamoindomaindynamicspretraining} adopted a similar inverse–forward dynamics approach to learn visual  representations. LAPA~\cite{ye2025latentactionpretrainingvideos} and Moto-GPT~\cite{chen_moto_2025} (Moto) extended this paradigm to internet-scale and robot datasets by pretraining vision–language models for robotic policy learning. IGOR~\cite{chen2024igorimagegoalrepresentationsatomic} learns a latent space by jointly training on human and robot demonstrations. UniVLA~\cite{bu_univla_2025} introduces task-centric latent tokens for structured action reasoning, while Amplify~\cite{collins_amplify_2025} leverages keypoint trajectories as supervision to learn transferable latent actions between humans and robots. Villa-X~\cite{chen_villax_2025} further enhances latent representations by incorporating proprioceptive feedback for richer dynamics. Recently, several world models~\cite{agibotworldcontributors2025agibotworldcolosseolargescale,gao_adaworld_2025,zhu2025unifiedworldmodelscoupling} have adopted latent action spaces as a compact interface between perception and control. Despite these advances, most latent action methods learn action abstractions from a \emph{single} observation modality (typically RGB), and thus the learned latents primarily capture appearance-driven dynamics. Our work differs by  learning latent action representations jointly from both \emph{RGB and depth}, yielding a unified latent action representation that enhances appearance driven dynamics with 3D geometry for more precise manipulation.

\subsection{Enhancing VLAs with Depth.} 
Recent works have explored incorporating multiple visual modalities into VLA frameworks to enhance spatial understanding, perception, and control. In particular, several approaches integrate 3D scene information~\cite{yuan_depthvla_2025,bhat_3d-cavla_2025,patratskiy2025spatialtracesenhancingvla,li2025qdepthvlaquantizeddepthprediction} to improve geometric reasoning for robot manipulation.
3D-VLA~\cite{zhen20243dvla3dvisionlanguageactiongenerative} builds a 3D-aware VLA using a 3D-based LLM for spatial reasoning. Other methods, such as 3D-CaVLA~\cite{bhat_3d-cavla_2025}, DepthVLA~\cite{yuan_depthvla_2025}, QDepth-VLA~\cite{li2025qdepthvlaquantizeddepthprediction}, Spatial traces~\cite{patratskiy2025spatialtracesenhancingvla} employ depth-maps as auxiliary supervision, either by projecting depth features into the language-model embedding space or by introducing explicit depth-prediction heads to guide action learning. 
SpatialVLA~\cite{qu_spatial-vla_2025} introduces Ego3D position encoding and adaptive action grids, enabling precise 3D grounding and reasoning in VLAs.

However, these approaches process depth at the pixel or feature level rather than integrating 3D cues into the action abstraction itself. Incorporating depth into \emph{latent action} representations therefore remains under-explored. A closely related work to ours is UniSkill~\cite{kim2025uniskillimitatinghumanvideos}, which learns embodiment-agnostic skill representations from unlabeled human videos. While UniSkill leverages both RGB and depth inputs to learn generic skill representations, it primarily focuses on learning a common embedding space for the modalities jointly.  
In contrast, \methodabbrev first preserves modality-specific RGB and depth representations and subsequently fuses and re-quantizes them to obtain a unified latent representation. This design enables VLAs to perform cross-modal latent pretraining that transfers complementary semantic cues from RGB and geometry-grounded structure from depth, thereby providing scalable and structured pretraining signals.


\begin{figure*}[t]
    \centering

    \begin{minipage}{\textwidth}
        \centering
        \scalebox{0.85}{
       \resizebox{!}{\height}{
        \includegraphics[width=\textwidth]{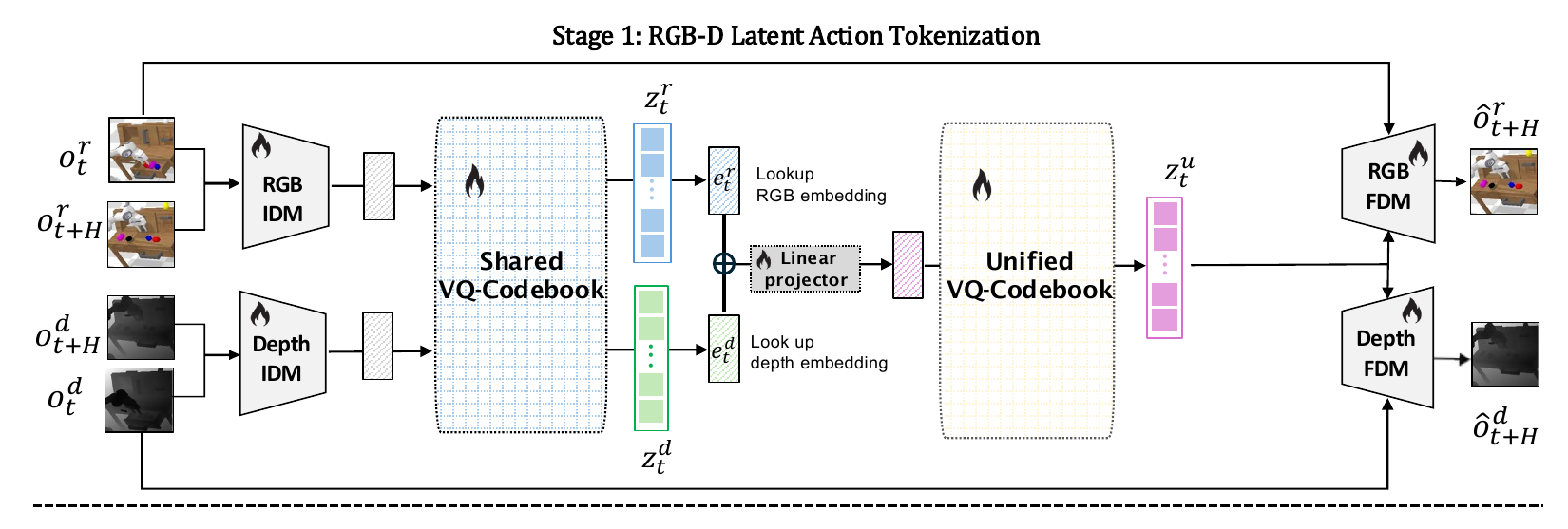}}}
    \end{minipage}
    \vspace{0.4em}
    \begin{minipage}{\textwidth}
        \centering
        \scalebox{0.85}{
         \resizebox{!}{\height}{
        \includegraphics[width=0.95\textwidth]{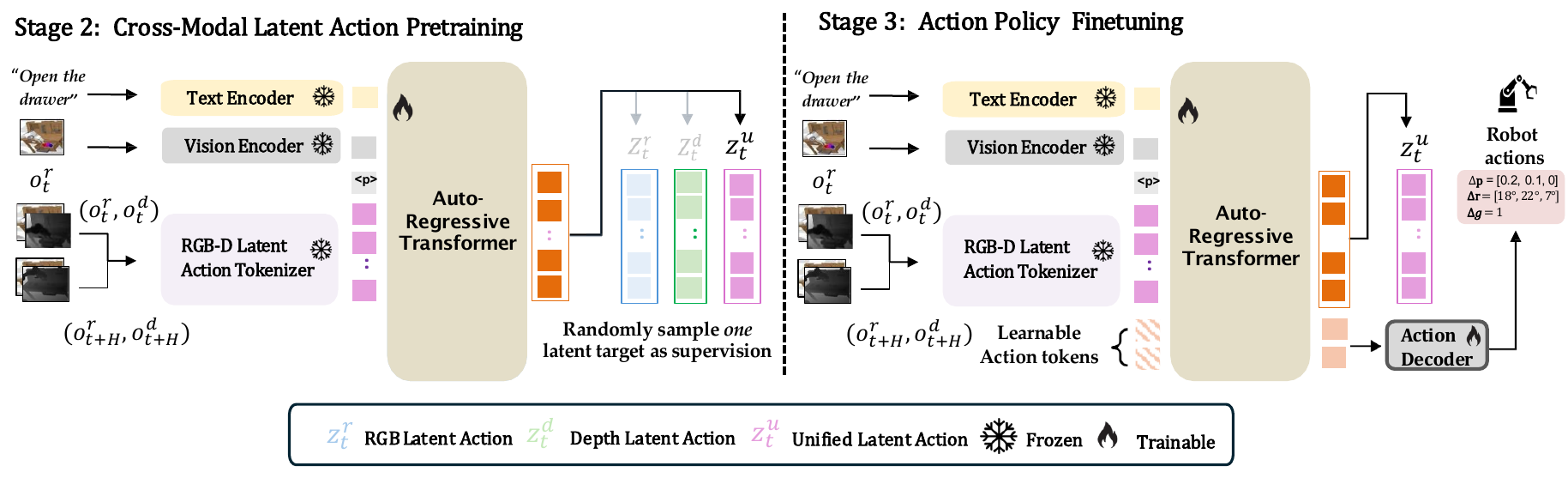}}}
    \end{minipage}

    \caption{\textbf{Overview of \methodabbrev three stages:} (1) We train an RGB-D Latent Action Tokenizer to obtain three discrete latent actions from pairs of RGB-D observations: \colorbox{rgbbox}{RGB}, \colorbox{depthbox}{Depth}, and \colorbox{unibox}{Unified} RGB-Depth. (2) \methodabbrev is pretrained with cross-modal autoregressive latent-token prediction to capture complementary priors from RGB appearance and depth geometry. (3) \methodabbrev is fine-tuned on action-labeled trajectories to map predicted latent actions to executable robot actions.}
    \label{fig:UniLACT_method}
\end{figure*}
\vspace{-2pt}
\section{Proposed Method}

RGB and depth provide complementary views of the same state transition as RGB captures appearance and semantic changes, while depth exposes the underlying geometric motion. However, simply fusing the two modalities can obscure modality-specific information, whereas learning them independently fails to model their correspondence. Our key idea is therefore to preserve RGB- and depth-specific supervision while simultaneously learning a unified action representation that captures what the two modalities share.

To this end, \methodabbrev explicitly maintains three complementary latent-action targets for every RGB-D transition: an RGB-specific latent, a depth-specific latent, and a unified latent. These targets provide dedicated supervision for appearance dynamics, geometric dynamics, and their shared action semantics, respectively.

As illustrated in Figure~\ref{fig:UniLACT_method}, training proceeds in three stages. First, we learn the three latent-action targets from RGB-D observation pairs using a two-stage quantization scheme. Second, \methodabbrev is pretrained to predict the RGB-specific, depth-specific, and unified targets as separate pseudo-actions, allowing each supervisory signal to shape the latent action space without requiring ground-truth robot actions. Finally, the pretrained model is fine-tuned on action-labeled robot demonstrations to produce executable controls. Depth is used only during training; at inference, \methodabbrev operates from RGB observations and task instructions alone.


\subsection{RGB-D Latent Action Tokenization}
\label{sec:stage1}

Given paired RGB and depth observations, latent action tokenization learns modality-specific latent action representations together with a unified latent action representation in a shared discrete space. To achieve this, the tokenizer captures latent action representations and their cross-modal relationships through a two-step vector-quantization pipeline. For each modality, an inverse dynamics model (IDM) first maps the paired observations to a continuous latent embedding, which is subsequently discretized using a shared vector-quantization operator. The resulting modality-specific discrete latents are then fused and re-quantized to produce the unified latent representation. This unified latent is then used to condition modality-specific forward dynamics models (FDMs), ensuring it retains sufficient information to predict future observations in each modality, thereby preserving complementary dynamics across modalities.


Formally, we consider paired observations $\big(o_t^{m}, o_{t+H}^{m}\big)$ corresponding to the current and future frames of modality $m$, where $H$ denotes the temporal offset and $m \in \{\mathrm{r}, \mathrm{d}\}$ indicates RGB or depth, respectively. For each modality $m$, an IDM $I_m$ maps the observation pair to a continuous latent embedding,

\begin{equation}
\tilde{\mathbf{z}}_t^{m} = I_m\!\big(o_t^{m},\, o_{t+H}^{m}\big)
\end{equation}

This embedding is discretized by a vector-quantization operator $\operatorname{VQ}$ with respect to a codebook $\mathcal{C}^{(s)}$ shared across both RGB and depth modalities. Thus, the modality-specific latent action representation is given by

\begin{align}
z_t^{m} &= \operatorname{VQ}\!\big(\tilde{\mathbf{z}}_t^{m};\, \mathcal{C}^{(s)}\big)
\end{align}
Subsequently, modality-specific codebook embeddings $\mathbf{e}_t^{m}$ are obtained by 
\begin{align}
\mathbf{e}_t^{m} &= \mathcal{C}^{(s)}\!\left[z_t^{m}\right]
\end{align}
which are then concatenated and linearly projected into a continuous unified latent space such that,
\begin{equation}
\tilde{\mathbf{e}}_t^{\mathrm{c}} = \big[\mathbf{e}_t^{\mathrm{r}};\, \mathbf{e}_t^{\mathrm{d}}\big], 
\qquad
\mathbf{h}_t = \mathbf{W}_f\, \tilde{\mathbf{e}}_t^{\mathrm{c}} + \mathbf{b}_f
\end{equation}
where $\mathbf{W}_f$ and $\mathbf{b}_f$ are the parameters of the linear projector.
Then, the projected representation $\mathbf{h}_t$ is discretized by a second vector-quantization operator with respect to another codebook $\mathcal{C}^{(u)}$,
\begin{equation}
z_t^{u} = \operatorname{VQ}\!\big(\mathbf{h}_t;\, \mathcal{C}^{(u)}\big)
\end{equation}


We use two-step quantization to preserve what each modality contributes before learning what they share. RGB and depth are first discretized into modality-specific latent actions within a shared codebook, then combined and re-quantized into a unified latent action. This provides dedicated RGB and depth supervision while avoiding the inconsistency of independent latent spaces and the information loss of early fusion. We ablate this design choice in Table~\ref{tab:vq_ablation}.

Finally, the unified latent action representation $z_t^{u}$ is used together with each modality's current observation $o_t^m$ to condition modality-specific FDM $F_m$ to reconstruct future observation as
\begin{equation}
\hat{o}^{\,m}_{t+H} = F_m\!\big(o_t^{m},\, z_t^{u}\big)
\end{equation}
This disentangled reconstruction objective across modality-specific domains enforces the unified representation $z_t^u$ to capture the complementary dynamics of each modality. We refer to this complete parameterized model, which takes RGB and depth observations as input to learn both modality-specific and unified latent representations, as the \textit{RGB-D latent action tokenizer}.

\subsection{Cross-Modal Latent Action Pretraining}
\label{sec:stage2}
\label{sec:pretraining}

We use the pretrained RGB-D latent-action tokenizer to extract three latent-action targets from each RGB-D transition: RGB-specific $z_t^{r}$, depth-specific $z_t^{d}$, and unified $z_t^{u}$. These discrete latent actions serve as supervision for the second-stage VLA pretraining.

Given the current visual observation $o_t$, task instruction $l$, and unified latent representation $z_t^{u}$, \methodabbrev is trained autoregressively to predict one of the three latent-action types. Specifically, for a target latent type $m \in \{r,d,u\}$, the model predicts a sequence
$z^{m}_{1:N}$, where $m=r$, $d$, and $u$ correspond to RGB-specific, depth-specific, and unified latent actions, respectively. Here, $N$ is the number of discrete tokens used to represent a single action transition from $o_t$ to $o_{t+H}$. Thus, the index $i$ refers to tokens within one latent action, not to different timesteps. 

Particularly, for a selected target latent type $m$, we optimize the standard autoregressive next-token prediction objective:
\vspace{-3pt}
\begin{equation}
\mathcal{L}_{\text{latent}}^{m}
=
-\sum_{i=1}^{N}
\log p_\theta\!\left(
z^{m}_{i}
\,\middle|\,
z^{u}_{1:i-1},\, o_t,\, l
\right)
\label{eq:ar_loss}
\end{equation}

During pretraining, the target latent type is alternated across batches. Formally, we sample
$m \sim \mathcal{U}\{r,d,u\}$ and optimize the corresponding latent-action loss:
\begin{equation}
\mathcal{L}_{\text{pretrain}}
=
\mathbb{E}_{m \sim \mathcal{U}\{r,d,u\}}
\left[
\mathcal{L}_{\text{latent}}^{m}
\right]
\label{eq:pretrain_loss}
\end{equation}

\subsection{Action Policy Fine-Tuning}
\label{sec:stage3}



Finally, we fine-tune the pretrained model on action-labeled robot demonstrations using a standard action prediction objective~\cite{ye2025latentactionpretrainingvideos,chen_moto_2025}. Learnable action query tokens are appended to the input sequence, and their corresponding transformer outputs are mapped through a lightweight decoder to 7-DoF end-effector commands. Continuous translation and rotation are trained with an \(L_1\) loss and the gripper state with binary cross-entropy, while retaining the unified latent-action objective during fine-tuning:


\begin{equation}
\mathcal{L}_{\text{action}}
=
\mathcal{L}_{\text{reg}}\!\left(\Delta\mathbf{p}\right)
+
\mathcal{L}_{\text{reg}}\!\left(\Delta\mathbf{r}\right)
+
\mathcal{L}_{\text{bce}}\!\left(g\right)
\label{eq:action_loss}
\end{equation}
To preserve the pretrained latent structure while learning a robot policy, we fine-tune \methodabbrev using the sum of the unified latent-action loss $\mathcal{L}_{\text{latent}}^{u}$ and action-prediction loss:
\begin{equation}
\mathcal{L}_{\text{finetune}}
=
\mathcal{L}_{\text{latent}}^{u}
+
\mathcal{L}_{\text{action}}
\label{eq:ft_loss}
\end{equation}
At \textbf{inference} time, \methodabbrev requires only RGB observations and task instructions as input. 

\section{Experiments}
\label{sec:exp}

In this section, we first describe the implementation details and experimental benchmarks. We then evaluate \methodabbrev in both simulation and real-world environments. Finally, we analyze the run-time complexity of our model and key design choices through a comprehensive ablation study.

\subsection{Implementation Details}

\methodabbrev is implemented in PyTorch and all  experiments are conducted on 4 NVIDIA H200 GPUs.

\noindent\textbf{RGB-D Latent Action Tokenizer} primarily consists of modality-specific IDMs, FDMs, and discretization modules. Each IDM pairs a frozen MAE-pretrained~\cite{mae} ViT-L encoder~\cite{dosovitskiy2021imageworth16x16words} with a spatio-temporal transformer. We append 8 learnable query embeddings to the patch features to capture inter-frame dynamics, and discretize the resulting outputs via a VQ-VAE objective~\cite{van2017neural} into 8 latent action tokens of dimension 32. Each FDM is a ViT-B decoder~\cite{dosovitskiy2021imageworth16x16words} trained with an MSE loss to reconstruct target observations.

\noindent\textbf{\methodabbrev}'s policy is built upon the GPT-2 causal transformer backbone~\cite{radford2019language}. We encode task instructions using a frozen T5 encoder~\cite{T5}, and visual inputs are processed using ViT-L~\cite{dosovitskiy2021imageworth16x16words} vision encoder. To predict latent actions and real robot actions, we employ a latent decoder and an action decoder. The latent decoder uses a linear layer to project hidden states into discrete latent actions. The action decoder consists of a 2-layer MLP, followed by separate linear layers that output continuous robot arm actions and a binary gripper command.

\subsection{CALVIN Experiments}

We first evaluate \methodabbrev on the CALVIN benchmark~\cite{mees2022calvinbenchmarklanguageconditionedpolicy}, a simulated benchmark designed for long-horizon, multi-task, language-conditioned  manipulation tasks.

\noindent\textbf{Training Data.} The CALVIN benchmark consists of four environments (A, B, C, and D) and 34 manipulation tasks, providing approximately 18K trajectories that include RGB-D observations and natural language annotations. We follow the challenging ABC$\rightarrow$D setting, where models are trained on data from environments A, B, and C, and evaluated in the unseen environment D to assess generalization.

For in-domain training, we adopt the same setup as~\cite{wu2023unleashinglargescalevideogenerative}, using only the language-annotated ABC splits for VLA pretraining to ensure fair comparison. 
We additionally report results with out-of-domain pretraining, where \methodabbrev is pretrained on a subset of the Open X-Embodiment (OXE) dataset~\cite{open_x_embodiment_rt_x_2023}.

\noindent\textbf{Evaluation.}
We follow the standard evaluation protocol from ~\cite{mees2022calvinbenchmarklanguageconditionedpolicy}. Each policy is evaluated over 1,000 episodes. In each episode, the robot needs to accomplish a sequence of five consecutive tasks. Each task is allowed up to a maximum of 360 action steps unless successfully completed earlier. Performance is reported using per-task success rates and the average sequence length (\textit{Avg. Len.}), defined as the average number of consecutive tasks  completed within each episode.

\begin{table*}[t]
  \caption{CALVIN ($ABC\rightarrow D$) Benchmark results. Performance metric is the average sequence length. Here `*' represents the results are reproduced with our training settings and `Static RGB',` Gripper RGB' represents the static camera and gripper camera view respectively. `Proprio.' denotes the robot proprioceptive state.}
  \label{tab:sota_calvin}
  \begin{center}
  \setlength{\tabcolsep}{5pt}
  \renewcommand{\arraystretch}{1.0}
  \begin{tabular}{@{}llcccccc@{}}
      \toprule
     \multirow{2}{*}{\textbf{Model}} & \multirow{2}{*}{\textbf{Observation input}} & \multicolumn{5}{c}{\textbf{Tasks Completed in a row}}  & \multirow{2}{*}{\textbf{Avg. Len $\uparrow$}} \\
      \cmidrule(lr){3-7}
      & & T1 & T2 & T3 & T4 & T5 &  \\
      \midrule
      \rowcolor{blue!10}
    \multicolumn{8}{c}{\textit{CALVIN (ABC) Pretraining}} \\
    Diffusion Policy~\cite{chi2024diffusionpolicyvisuomotorpolicy} & Static RGB & 0.402 & 0.123 & 0.026 & 0.008 & 0.00 & 0.56 \\
    HULC~\cite{mees_what_2022} & Static RGB & 0.418 & 0.165 & 0.057 & 0.019 & 0.011 & 0.67 \\
    RT-1~\cite{brohan2023rt1roboticstransformerrealworld} & Static RGB & 0.533 & 0.222 & 0.094 & 0.038 & 0.013 & 0.90 \\
    Robo-Flamingo~\cite{li_roboflamingo_2024} & (Static + Gripper) RGB & 0.824 & 0.619 & 0.466 & 0.331 & 0.235 & 2.48 \\
    SuSIE~\cite{black2023zeroshotroboticmanipulationpretrained} & Static RGB & 0.870 & 0.690 & 0.490 & 0.380 & 0.260 & 2.69 \\
    \rowcolor{gray!10}
    GR-1~\cite{wu2023unleashinglargescalevideogenerative} & (Static + Gripper) RGB + Proprio. & 0.854 & 0.712 & 0.596 & 0.497 & 0.401 & 3.06 \\
    UniSkill~\cite{kim2025uniskillimitatinghumanvideos}* & Static (RGB + Depth) & 0.772 & 0.567 & 0.421 & 0.291 & 0.204 & 2.25 \\
    Moto~\cite{chen_moto_2025}* & Static RGB & 0.827 & 0.637 & 0.485 & 0.374 & 0.278 & 2.60 \\
    \textbf{\methodabbrev (Ours)} & Static RGB & \textbf{0.855} & \textbf{0.689} & \textbf{0.543} & \textbf{0.440} & \textbf{0.332} & \textbf{2.86} \\
    \midrule
    \rowcolor{blue!10}
    \multicolumn{8}{c}{\textit{OXE Pretraining} + \textit{CALVIN (ABC) Finetuning}} \\
    Moto~\cite{chen_moto_2025}* & Static RGB & 0.874 & 0.682 & 0.519 & 0.396 & 0.298 & 2.77 \\
    \textbf{\methodabbrev (Ours)} & Static RGB & \textbf{0.897} & \textbf{0.729} & \textbf{0.584} & \textbf{0.493} & \textbf{0.401} & \textbf{3.10} \\
    \bottomrule
  \end{tabular}
  \end{center}
  \vspace{-5pt}
\end{table*}

\noindent\textbf{Results.}
On the CALVIN ABC$\rightarrow$D benchmark, policy learning approaches broadly fall into several categories, including (1) language-conditioned behavioral cloning, (2) diffusion-based policies, and (3) planner–executor frameworks. Our method belongs to the first category; therefore, Table~\ref{tab:sota_calvin} compares against representative baselines within this group. We report results under both in-domain and out-of-domain pretraining settings. 

For \underline{in-domain} results, since we follow the evaluation protocol of prior work~\cite{wu2023unleashinglargescalevideogenerative, chen_moto_2025}, baseline numbers are directly taken from these references. We reproduce results of Moto~\cite{chen_moto_2025}, an RGB latent-action-based approach most closely related to our method. Under comparable settings, \methodabbrev consistently outperforms Moto, demonstrating the benefit of incorporating depth into the latent action representation. We next compare against UniSkill~\cite{kim2025uniskillimitatinghumanvideos}, implemented in our setting, which underperforms both the RGB-only Moto and \methodabbrev, as combining both modalities naively lets one modality dominate and suppress the other. \methodabbrev instead retains modality-specific latents alongside the unified latent, preserving each modality's contribution. We also observe that \methodabbrev surpasses baselines operating with comparable input modalities. Although GR-1 reports the strongest overall performance, it relies on additional inputs such as gripper-view RGB images and proprioceptive state, whereas \methodabbrev uses only static RGB observations.

For \underline{out-of-domain} pretraining on OXE, \methodabbrev outperforms Moto by \textcolor{gaingreen}{\textbf{+11.9\%}} in average sequence length, confirming that large-scale pretraining improves generalization. Notably, this gain stems from more than data scale alone; while scaling from CALVIN to OXE boosts the RGB-only baseline, \methodabbrev improves further through geometric information in the unified latent and cross-modal supervision from the modality-specific latents, indicating that the gain is driven by the mechanism rather than data scale.

\begin{figure}[h]
    \centering
    \setlength{\abovecaptionskip}{2pt}
    \setlength{\belowcaptionskip}{0pt}
    \includegraphics[height=5in, width=\linewidth, keepaspectratio]{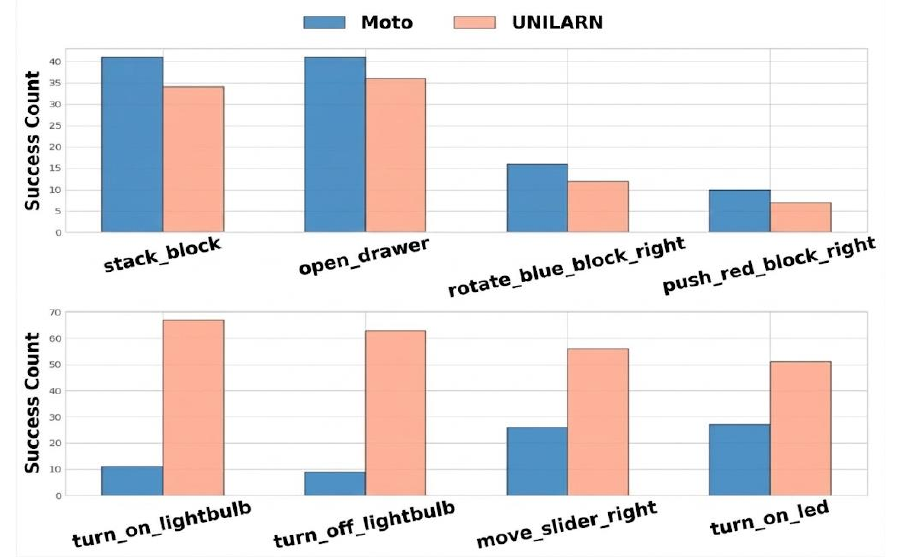}
    \caption{\textbf{Task-wise CALVIN success comparison.} \textbf{Top:} tasks favoring RGB latents; \textbf{Bottom:} tasks favoring unified RGB-D latents.}
    \label{fig:rgb_depth_quantitative}

    \vspace{4pt}
    \includegraphics[height=1.2in, width=0.75\linewidth]{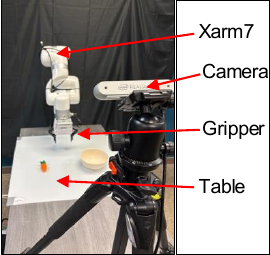}
    \caption{\textbf{Real-world setup.} xArm7 manipulator with a two-fingered parallel gripper and a table-mounted workspace camera.}
    \label{fig:xarm7_setup}
    \vspace{-15pt}
\end{figure}

To better understand the role of depth, we further analyze task-level performance on CALVIN (Figure~\ref{fig:rgb_depth_quantitative}). RGB-based latent representations perform competitively on appearance-driven tasks (e.g., \textit{stack blocks}, \textit{rotate blue block}), whereas unified (RGB+depth) latents yield larger gains on geometry-centric tasks (e.g., \textit{move slider}, \textit{turn on light bulb}). These results suggest that incorporating depth enriches the latent action space with stronger geometric priors, which are critical for precise and contact-rich manipulation.


\subsection{Real-World Experiments}

We further evaluate \methodabbrev on a 7-DoF xArm7 manipulator using third-person RGB-D observations from a front-mounted Intel RealSense D435i camera, shown in Figure~\ref{fig:xarm7_setup}. We collect 30 teleoperated demonstrations~\cite{khazatsky2025droidlargescaleinthewildrobot} per task and evaluate policies across five tabletop manipulation tasks: 

\begin{itemize}\setlength{\itemsep}{1pt}\setlength{\parskip}{0pt}
\item \texttt{T1: Put carrot in the bowl.} Pick up the carrot and place it completely inside the bowl.
\item \texttt{T2: Move \textless object1\textgreater~near \textless object2\textgreater.} Move \textless object1\textgreater~within a 3~cm radius of \textless object2\textgreater, with both randomly sampled from eight toys: \textit{\{eggplant, lemon, banana, peach, carrot, green pepper, strawberry, tomato\}}.
\item \texttt{T3: Stacking cubes.} Pick up the blue cube, initialized at a random horizontal position, and stack it stably on the green cube.
\item \texttt{T4: Put the eggplant into the plate.} Pick up the eggplant and place it inside the plate.
\item \texttt{T5: Lift the block.} Grasp the block and lift it more than 1~cm above the table.
\end{itemize}
Tasks T1--T3 are \emph{seen} during training, whereas T4 and T5 are \emph{unseen} tasks used to evaluate zero-shot generalization.

\noindent\textbf{Training Data.}
\methodabbrev is trained on a subset of the OXE dataset~\cite{open_x_embodiment_rt_x_2023} together with collected real-world demonstrations. The model is subsequently fine-tuned using only real-world demonstrations. Since most of the OXE subset does not provide depth observations, we generate depth maps from RGB images using an off-the-shelf depth estimation foundation model~\cite{yang2024depthv2}.

\noindent\textbf{Evaluation.} Each task is evaluated over 20 trials with randomized, out-of-distribution initial object configurations. We report per-task and average success rates across all tasks.

\begin{table}[t]
\centering
\setlength{\abovecaptionskip}{2pt}
\setlength{\belowcaptionskip}{2pt}
\caption{\textbf{Real-world results.} Success rate (\%) on seen and unseen tasks.}
\label{tab:real-robot-results}
\footnotesize
\setlength{\tabcolsep}{6.5pt}
\renewcommand{\arraystretch}{0.88}
\setlength{\aboverulesep}{1.5pt}
\setlength{\belowrulesep}{1.5pt}
\begin{tabular}{@{}lcccccc@{}}
\toprule
& \multicolumn{3}{c}{\textbf{Seen}} & \multicolumn{2}{c}{\textbf{Unseen}} & \\
\cmidrule(lr){2-4} \cmidrule(lr){5-6}
\textbf{Method} & T1 & T2 & T3 & T4 & T5 & \textbf{Avg.} \\
\midrule
Moto~\cite{chen_moto_2025} & 55 & 40 & 25 & 45 & 30 & 39 \\
\textbf{\methodabbrev} & \textbf{75} & \textbf{75} & \textbf{65} & \textbf{50} & \textbf{40} & \textbf{61} \\
\bottomrule
\end{tabular}
\vspace{-10pt}
\end{table}

\begin{figure}[!h]

    \centering
        \centering
             \includegraphics[width=\linewidth,height=3cm]{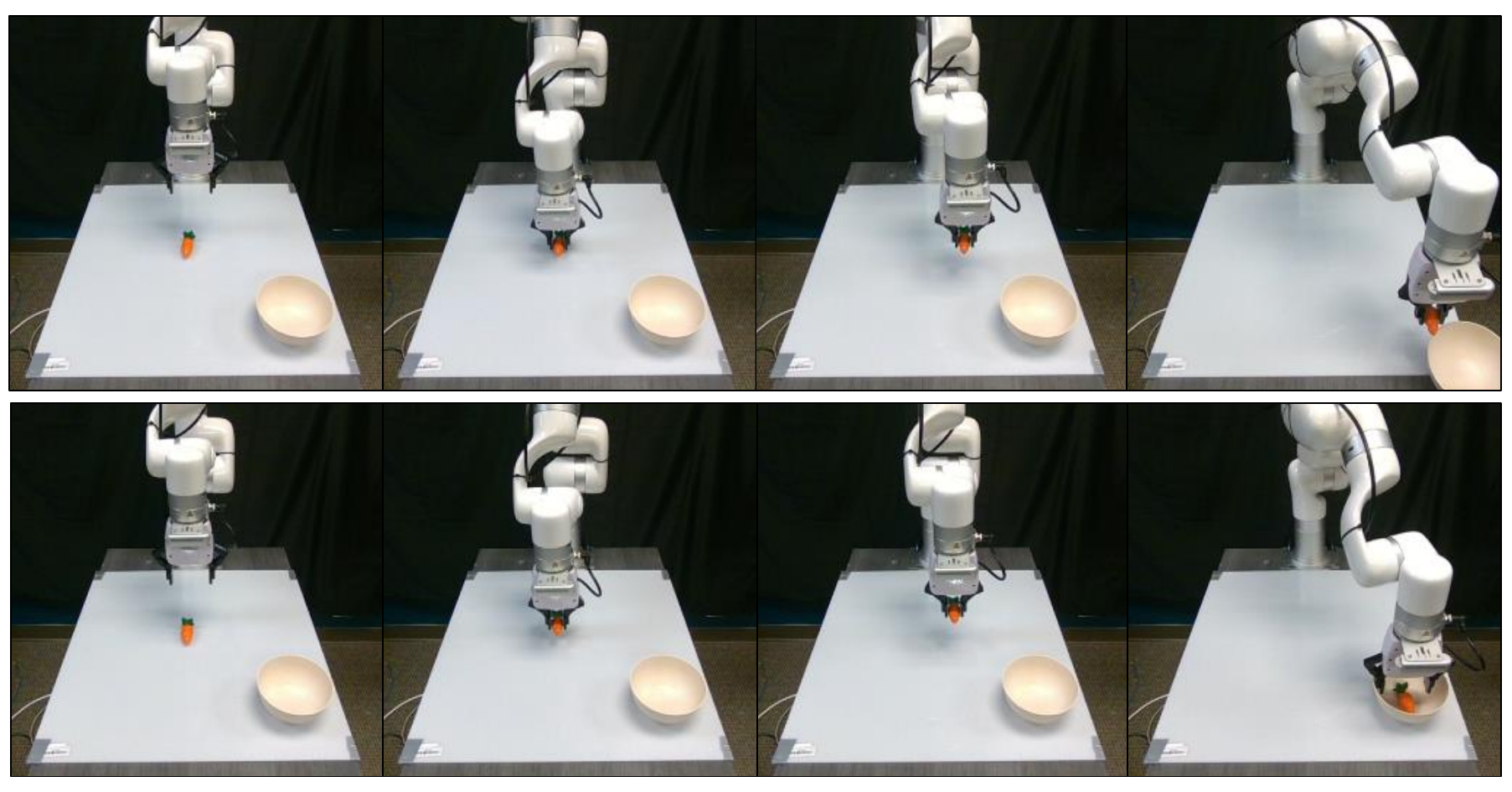}
       \caption{\textbf{Illustration of Task T1 execution comparing Moto and \methodabbrev.}  \textbf{Top row:} Moto fails to place the carrot inside the bowl and pushes the bowl out of the workspace. 
\textbf{Bottom row:} \methodabbrev successfully completes the task.}
        \label{fig:Task_1_Viz}
\end{figure}

\begin{figure}[!h]
        \centering
        \includegraphics[width=\linewidth,height=3cm]{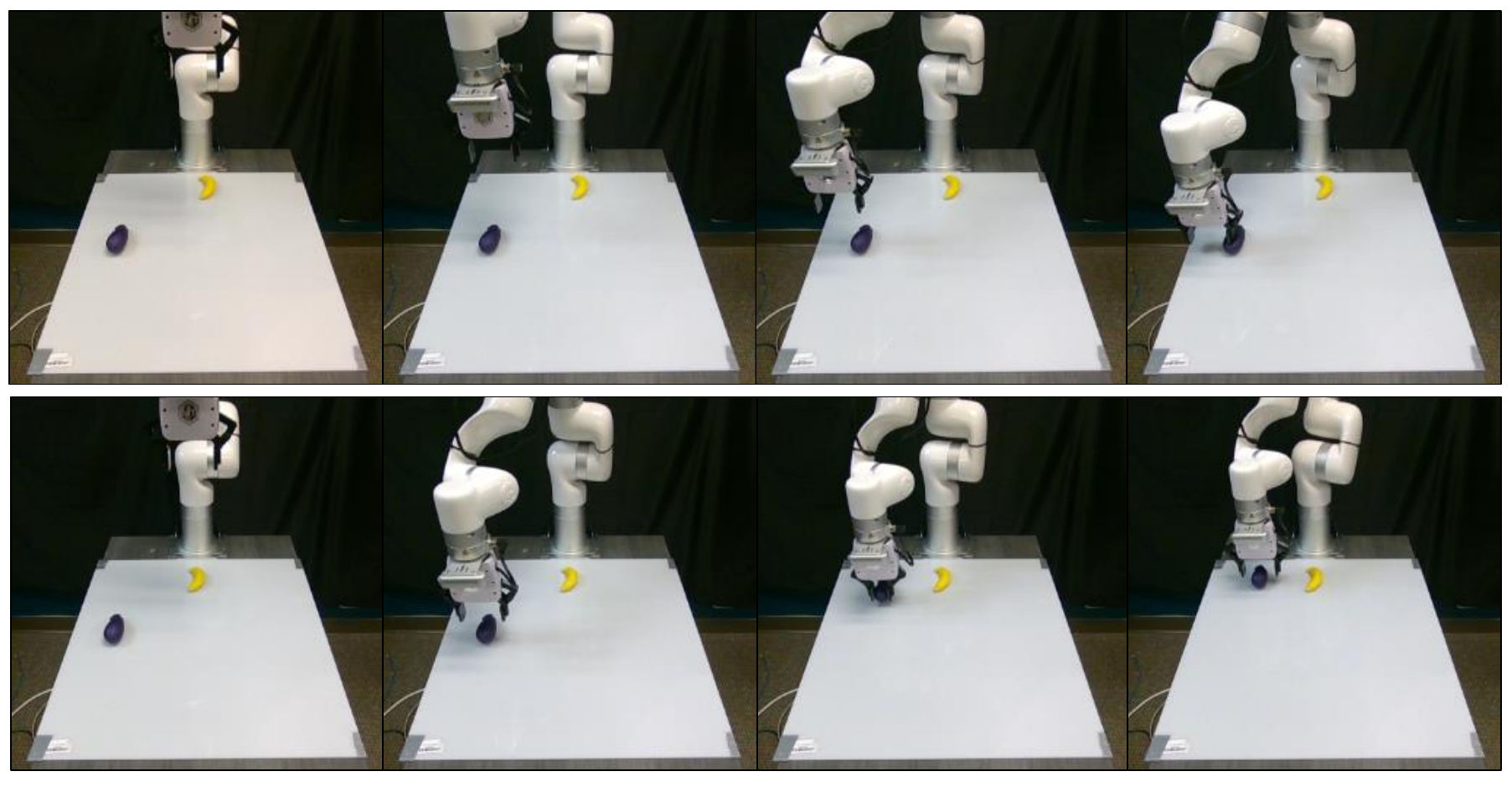}
     \caption{\textbf{Illustration of Task T2:} ``Move the eggplant near the banana.'' 
\textbf{Top row:} Moto approaches the eggplant but fails to grasp it and collides with the workspace. 
\textbf{Bottom row:} \methodabbrev successfully grasps the eggplant and moves it near the banana.}
    \label{fig:Task_2_Viz}
\vspace{-10pt}
\end{figure}

\noindent\textbf{Results.}
Table~\ref{tab:real-robot-results} shows that \methodabbrev~ improves over the RGB-latent baseline, Moto, on all tasks, with a \textcolor{gaingreen}{\textbf{56.4\%}} improvement in average success rate. The gains are concentrated on the seen tasks (T1, T2, T3); on the two unseen tasks, \methodabbrev and the RGB-latent baseline perform comparably, indicating that the depth-aware latents primarily benefit tasks seen during training while generalizing on par with the baseline. This performance improvement indicates that the geometric information encoded in \methodabbrev's latent representations significantly enhances the policy's 3D spatial understanding. We also provide a qualitative comparison in Figures~\ref{fig:Task_1_Viz} and \ref{fig:Task_2_Viz}, which demonstrate how precisely \methodabbrev accomplishes the tasks while avoiding collision. In contrast, RGB-latent-based Moto inaccurately estimates the height of the objects, leading to collision . Overall, real-world results show that \methodabbrev enhances the depth awareness of latent action representations, allowing for precise manipulation. The significant performance gains in both simulation and real-world environments suggest that \methodabbrev better interprets the environment's 3D structure than methods that rely on appearance-only latents.

\vspace{-0.5em}

\subsection{Computational Analysis}

We compare the computational complexity of \methodabbrev with the RGB-based latent action  model (Moto). Table~\ref{tab:model_complexity} reports the number of model parameters and the per-step inference latency. 
 Despite incorporating both RGB and depth information into a unified latent space during training, \methodabbrev maintains the same model size and runtime as Moto. This is enabled by the design of \methodabbrev, which requires the additional depth modality only during training and does not use it during inference.

\begin{table}[h]
\centering
\scriptsize
\setlength{\tabcolsep}{6pt}
\renewcommand{\arraystretch}{1.05}
\caption{Deployment complexity}
\label{tab:model_complexity}
\begin{tabular}{@{}lcc@{}}
\toprule
\textbf{Model} & \textbf{\#P (M)} & \textbf{Lat.\ (ms)} \\
\midrule
Moto & 89.8 & 27 \\
\methodabbrev & 89.8 & 27 \\
\bottomrule
\end{tabular}
\end{table}

\subsection{Depth Quality Analysis: Estimated versus Native Depth}
We investigate the effect of estimated-depth quality on our method by first quantifying its deviation from native depth on CALVIN and then evaluating its impact on downstream policy performance. Since depth annotations are unavailable for most of OXE and are therefore estimated rather than measured, we assess whether pseudo-depth obtained from a monocular depth estimator~\cite{yang2024depthv2} can serve as an effective substitute for native sensor depth. On CALVIN, where native depth is available, the estimated depth achieves an AbsRel of $0.044$, indicating close agreement with the sensor measurements (Table~\ref{tab:depth_sensitivity}). To determine whether this level of fidelity is sufficient for policy learning, we perform OXE pretraining and subsequently fine-tune on CALVIN using latent-action targets computed from either pseudo-depth or native depth. We further probe the frozen latent representations for encoded depth information (Table~\ref{tab:depth_probing}). Under both native and estimated depth, the unified latent achieves a higher $R^2$ than the RGB-only latent, indicating that the unified representation retains modestly more geometric information.

\vspace{-5pt}
\begin{table}[t]
\centering
\setlength{\tabcolsep}{4pt}
\footnotesize
\begin{minipage}[t]{0.48\columnwidth}\centering
\caption{Depth input sensitivity}
\label{tab:depth_sensitivity}
\begin{tabular}{@{}lcc@{}}
\toprule
Depth source & AbsRel$\downarrow$ & Avg.\ Len.$\uparrow$ \\
\midrule
Pseudo (DA-V2) & 0.044 &  3.08 \\
Native         & 0.000 & 3.10\\
\bottomrule
\end{tabular}
\end{minipage}
\hfill
\begin{minipage}[t]{0.48\columnwidth}\centering
\caption{Depth probing (R$^2\uparrow$)}
\label{tab:depth_probing}
\begin{tabular}{@{}lccc@{}}
\toprule
Latent & Native & DA-V2 \\
\midrule
RGB-only & 0.705 & 0.488 \\
Unified  & \textbf{0.714} & \textbf{0.510} \\
\bottomrule
\end{tabular}
\end{minipage}

\end{table}

\subsection{Ablation Studies}

We conduct ablation studies on the CALVIN ABC\(\rightarrow\)D benchmark. We report the average sequence length to assess (a) the design choice of the RGB-D latent action tokenizer, (b) the effectiveness of various visual modalities, (c) the impact of various output modalities using unified latent actions while pretraining and fine-tuning, and (d) the impact of multi-task and single-task learning.

\begin{table}[t]
\centering
\scriptsize
\setlength{\abovecaptionskip}{2pt}
\setlength{\belowcaptionskip}{2pt}
\setlength{\tabcolsep}{4.5pt}
\renewcommand{\arraystretch}{0.9}
\caption{Ablation on design choices for the RGB-D tokenizer.}
\label{tab:vq_ablation}
\begin{tabular}{@{}lcccc@{}}
\toprule
\textbf{Variant} & \multicolumn{2}{c}{\textbf{Codebook}} & \textbf{Fused} & \textbf{Avg.\ Len.}$\uparrow$ \\
\cmidrule(lr){2-3}
& $\mathcal{C}^{(s)}$ & $\mathcal{C}^{(u)}$ & & \\
\midrule
Independent & \cmark & \xmark & -- & 2.080 \\
Fuse-once & \xmark & \cmark & Cont. & 2.254 \\
Per-modality & \cmark & \xmark & Disc. & 2.057 \\
\textbf{Two-step} & \cmark & \cmark & Disc. & \textbf{2.859} \\
\bottomrule
\end{tabular}
\vspace{-8pt}
\end{table}

\noindent\textbf{Design Choice of RGB-D Latent Action Tokenization.} Table~\ref{tab:vq_ablation} ablates our quantization design under identical RGB-D inputs, backbones, and training settings. \textit{Independent} quantization without cross-modal fusion introduces representation inconsistency across sensor streams, yielding 2.080 average sequence length. \textit{Fuse-once} quantizes continuous fused features directly via $\mathcal{C}^{(u)}$, discarding discrete modality-specific structure and letting high-variance RGB signals dominate depth geometry. \textit{Per-modality} fuses discrete tokens from $\mathcal{C}^{(s)}$ but omits $\mathcal{C}^{(u)}$, limiting cross-modal coordination during complex manipulation. Our two-step design resolves these failure modes by first quantizing each modality through a shared codebook $\mathcal{C}^{(s)}$ and subsequently compressing fused discrete tokens via a unified codebook $\mathcal{C}^{(u)}$. Crucially, retaining discrete modality-specific tokens also enables pretraining on RGB-only datasets, a capability that single-codebook approaches cannot support.

\begin{table}[t]
\centering
\small
\setlength{\tabcolsep}{6pt}
\renewcommand{\arraystretch}{0.6}
\caption{Impact of visual input modality.}
\label{tab:ablation_visual_modality}
\begin{tabular}{@{}lc@{}}
\toprule
\textbf{Latent Modality} & \textbf{Avg. Len}$\uparrow$ \\
\midrule
No latent actions & 0.744 \\
RGB & 2.601 \\
Depth & 2.402 \\
\midrule
RGB+Depth & 2.080 \\
Unified & 2.254 \\
\textbf{RGB+Depth+Unified} & \textbf{2.859} \\
\bottomrule
\end{tabular}
\vspace{-0.5em}
\end{table}

\noindent\textbf{Impact of visual modality.} Table~\ref{tab:ablation_visual_modality} analyzes how latent actions from different visual modalities contribute to policy learning. Both RGB and depth-based latent actions significantly outperform the no-latent baseline, confirming that pretraining provides essential complementary motion priors. We further compared \methodabbrev against a naive multimodal approach where RGB and depth latent actions are learned independently. In this setup, the model is trained with alternating supervision, where individual training batches are supervised by either RGB or depth latent actions. Although both modalities are used, learning modality-specific latents independently leads to representation inconsistency, resulting in a performance drop compared to individual modalities. We also evaluate a UniSkill-style variant (\textit{Unified}) that learns only a joint latent space without modality-specific latents. While \textit{Unified} improves over \textit{RGB+Depth}, the best results are achieved by \methodabbrev (Unified + Modality-specific), which jointly aligns RGB and depth in a shared latent space while retaining modality-specific latents.

\begin{table}[t]
\centering
\scriptsize
\setlength{\abovecaptionskip}{2pt}
\setlength{\belowcaptionskip}{2pt}
\renewcommand{\arraystretch}{0.95}
\begin{minipage}[t]{0.58\columnwidth}
\centering
\setlength{\tabcolsep}{2.5pt}
\caption{Output modalities ($\{U,R,D\}$ = Unified, RGB, depth).}
\label{tab:ablation_input_output}
\begin{tabular}{@{}lcc@{}}
\toprule
\textbf{Pretrain} & \textbf{Finetune} & \textbf{Avg.\ Len.}$\uparrow$ \\
\midrule
$U\!\rightarrow\!\{U,R,D\}$ & $U\!\rightarrow\!\{U,R,D\}$ & 2.299 \\
$U\!\rightarrow\!U$ & $U\!\rightarrow\!U$ & 2.597 \\
$U\!\rightarrow\!\{U,R,D\}$ & $U\!\rightarrow\!U$ & \textbf{2.859} \\
\bottomrule
\end{tabular}
\end{minipage}%
\hfill
\begin{minipage}[t]{0.39\columnwidth}
\centering
\setlength{\tabcolsep}{3pt}
\caption{Pretraining objectives.}
\label{tab:multi_vs_single_task}
\begin{tabular}{@{}lc@{}}
\toprule
\textbf{Variant} & \textbf{Avg.\ Len.}$\uparrow$ \\
\midrule
Multi-task & 2.055 \\
\textbf{Single-task} & \textbf{2.859} \\
\bottomrule
\end{tabular}
\end{minipage}
\vspace{-8pt}
\end{table}

\noindent\textbf{Impact of different output modalities.}
Next, we study whether pretraining and fine-tuning should predict all three latent types (RGB, depth, and unified) or only the unified latent. Table~\ref{tab:ablation_input_output} shows that pretraining with cross-modal targets (\( \mathrm{U}\!\rightarrow\!\{\mathrm{U},\mathrm{R},\mathrm{D}\} \)) and fine-tuning only on the unified latent (\( \mathrm{U}\!\rightarrow\!\mathrm{U} \)) yield the best results. This suggests that cross-modal supervision enriches the latent space during pretraining, while a compact unified target at fine-tuning leads to more effective action prediction. For fair comparison, all models are trained for a fixed number of iterations.

\noindent\textbf{Impact of training objectives.}
We compare single-task and multi-task training settings (Table~\ref{tab:multi_vs_single_task}). In both cases, the input to the model is the unified latent. In the multi-task setting, the model jointly predicts RGB, depth, and unified latents, whereas in the single-task setting it predicts one modality at a time. The single-task variant achieves higher performance, suggesting that jointly optimizing multiple latent objectives may introduce task interference, leading to less effective optimization.

\section{Conclusion}
We present \methodabbrev, a depth-aware latent action pretraining framework that brings 3D geometric awareness to vision-language-action models without requiring robot action annotations. Specifically, \methodabbrev extracts RGB-specific, depth-specific, and unified latent actions from RGB-D observations as pseudo-action targets for cross-modal policy pretraining. Experiments in simulation and on a real robot confirm the benefit of incorporating depth into the latent action space, yielding the greatest gains on geometry-centric tasks while requiring only RGB inputs at deployment. Crucially, preserving modality-specific dynamics alongside a shared unified latent outperforms collapsing modalities into a single space, highlighting the value of structured 3D spatial representations for robot policy learning.

\section*{ACKNOWLEDGMENT}
This work was supported in part by the National Science Foundation (IIS-2245652) and the University of North Carolina at Charlotte. Computational resources were provided by the NSF National AI Research Resource Pilot (NAIRR240338) and NCShare. We thank Nitin Chandrasekhar and  Smit Patel for collecting the teleoperation trajectories. We would also like to thank Xiang Li for our valuable discussions.






{\scriptsize
\renewcommand{\baselinestretch}{0.93}\selectfont
\setlength{\itemsep}{0pt}
\setlength{\parskip}{0pt}
\setlength{\parsep}{0pt}
\bibliographystyle{IEEEtran}
\bibliography{main}
}

\end{document}